# Knowledge Embedding and Retrieval Strategies in an Informledge System

Dr T.R. Gopalakrishnan Nair
Bio Computing Research Group
(Director): Research & Industry Incubation Center
D. S. Institutions
Bangalore, India
trgnair@yahoo.com,trgnair@ieee.org

Meenakshi Malhotra
Bio Computing Research Group
(Associate Member): Research & Industry Incubation Center
D. S. Institutions
Bangalore, India
uppal_meenakshi@yahoo.co.in

*Abstract* — Informledge System (ILS) is a knowledge network with autonomous nodes and intelligent links that integrate and structure the pieces of knowledge. In this paper, we put forward the strategies for knowledge embedding and retrieval in an ILS. ILS is a powerful knowledge network system dealing with logical storage and connectivity of information units to form knowledge using autonomous nodes and multi-lateral links. In ILS, the autonomous nodes known as Knowledge Network Nodes ( KNN)s play vital roles which are not only used in storage, parsing and in forming the multi-lateral linkages between knowledge points but also in helping the realization of intelligent retrieval of linked information units in the form of knowledge. Knowledge built in to the ILS forms the shape of sphere. The intelligence incorporated into the links of a KNN helps in retrieving various knowledge threads from a specific set of KNNs. A developed entity of information realized through KNN forms in to the shape of a knowledge cone.

## I. INTRODUCTION

Research in the field of Artificial Intelligence and Cognitive systems has been active for the last few decades to represent knowledge systematically. Knowledge bases have been used to store information based on domain ontologies. According to Dong-hui Yuan1, 2, Da-you Liu1, 2, Shi-qun Shen1, 2, Pu Yan1, 2[1], the semantic retrieval by search engines have been able to provide the result matching for the accurate query only. Also, they could not effectively model relationships between concepts and could not model events that change with time.

To overcome this, we had suggested an Informledge System (ILS) in our work published earlier [2]. ILS is a modified knowledge network that transforms information stored into meaningful knowledge by virtue of its autonomous nodes and multi-lateral links [2]. In ILS, we represent anything that exists as information, using autonomous nodes, even the synonyms which represent similar concepts, are represented separately with the help of links, as their common usage does not necessarily denote semantically equivalent concepts. The conventional domain ontologies result into conflicting term where the same term maybe chosen by two systems to denote completely different concepts [3]. For example, apple is used to denote fruit or computer. ILS represent both using the link properties that apple is a fruit and a computer is made by a company named 'Apple'. Many times the existing ontologies fail to differentiate between relationships and concentrates on words with preconceived connectivity requirements,

This paper is organized as follows. Section II discusses Research Background. We present our Research Objective in section III. Section IV gives the Conceptual Model of the Informledge System and in section V we answer why ILS is a knowledge network system. In Section VI we discuss how knowledge embedding and retrieval is done in ILS and section VII details about system simulation and evaluation. Finally, we conclude in section VIII.

## II. RESEARCH BACKGROUND

Although researchers in Cognitive and Artificial Intelligence field have been fantasizing to simulate the knowledge of human brain, but not much is achieved. While creating knowledge bases emphasis has been on the word and sentences, which are part of information sharing, rather than the concepts.

Ontologies have been used to form knowledge bases [4]. Ontology finds its origin in the field of philosophy and is understood as specification of what exists [5]. However, it's not always possible to represent all kinds of knowledge into ontology appropriate formats [6].

Knowledge bases are typically structured into a taxonomy tree where each node represents a concept and each concept has its parent as general concepts [7]. However, the definition of concept have been limited to just a word in all these knowledge bases. In ILS we consider concept as a fairly ordered set of entities. A knowledge thread can be a one or combination of these entities, that is it may be just a part of concept or it can denote a whole concept.

Even human brain uses prior knowledge for decision making, through the use of concepts which are formed through abstraction and capture the shared meaning of similar entities [8]. ILS does not deal with language directly but represent the idea or the thought. The same has also been accepted by the field of neuroscience. They also state that the different regions of the brain are responsible for concept formation and representation into language [9]. Which implies concept formation is independent of the language used, even people who cannot speak or infants with no knowledge of language can form concepts. Also as people belonging to different lingual are able to think of a concept and then are able to convey their thoughts in their respective languages. Language is just a representation of the knowledge which is formed by linking different knowledge points. As of now ILS deals with the linking of the knowledge points and not the representation part. However, language is used here to show which knowledge units are linked together.

## III. RESEARCH OBJECTIVE

Current ontological approach does not sufficiently meet the requirement of intelligently connecting information points together. So there is a need to realize a breakthrough in configuring an architecture where the nodes and links are sufficiently intelligent to store, interpret and form linkages, to create a knowledge thread from its component knowledge. Thus inter domain knowledge can be integrated without any differences in representation.

Mimicking the brain is one of the fascinating approaches followed by many people to realize knowledge capabilities. There have been attempts to reverse-engineer the mammalian brain as done by IBM Blue Brain Project [10]. Blue Brain Project mainly concentrates on how the brain functions so that they can assist in the field of neuro science.

Wherein ILS focuses on how information can be stored in a computing system that can simulate the knowledge system, which can be retrieved systematically as required. ILS proposed earlier [2] provides an opportunity to embed knowledge in a network of intelligent nodes which are autonomous and connect components of knowledge through intelligent ways. In this paper we further advance our work through implementation of a sample space of knowledge in a pragmatic Informledge unit.

In the following sections we see what ILS consists of and the role KNN plays in knowledge embedding and retrieval.

## IV. INFORMLEDGE SYSTEM

ILS is an autonomous knowledge network system comprising of intelligent nodes, storage and connectivity of information nodes to form knowledge. The knowledge in ILS depends on KNNs and multi-lateral links.

### A. Knowledge Network Node

ILS has intelligent nodes known as Knowledge Network Nodes (KNN). The structure of KNN as discussed in our earlier paper [2] is shown in Fig.1. The nodes in the ILS are homogeneous that is, nodes belonging to different domains are represented by same structure making the system uniform and scalable. As a result, every concept in ILS is represented using KNN so the system has uniform representation for all domains and sub-domain.

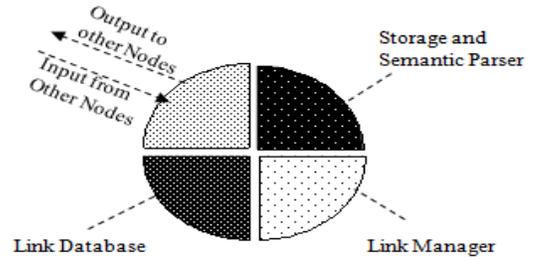

Figure 1  Knowledge Network Node

Concepts which are made from KNNs belonging to different domains can be linked together cohesively. KNN can not only store information but can have an inbuilt parser which has the ability to infer and reason the linkages to different nodes.

### B. Multi-lateral Links

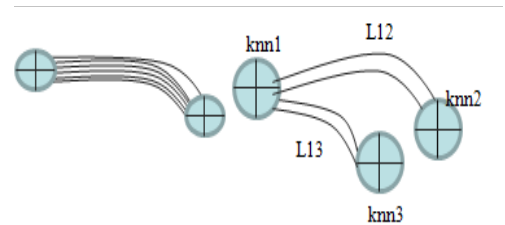

Figure 2  Links with Multiple Strands

Links are the most important part of ILS. Links are composed of multiple strands, where each strand is holding a property, as shown in Fig 2.

Rules have to be defined how the two nodes are linked. No two nodes can get arbitrarily linked. There have to be some associative property that can link the nodes. A thread of knowledge is created by connecting KNNs through links and the links are formed by combining the properties of all strands of a link.

TABLE 1    LINK PROPERTIES

| Property | Type |
| --- | --- |
| $α_{p1}$ | Directional (source/destination-KNN) |
| $α_{p2}$<br>$α_{p3}$<br>$α_{p4}$ | Performance<br>-Exclusive/Inclusive<br>-Additive/Subtractive<br>-Integrative/Differentiate |

Each link will have different features at different times when creating knowledge. Different properties of a link are listed in table 1 given below. Thus with these properties as shown in table 1, we are able to link the knowledge components to embed the knowledge which could be retrieved automatically.

During knowledge retrieval self-processing links are created based upon link properties between two KNNs

$$\alpha_{L12} = (\alpha_{p11,2}+\alpha_{p12,2}+\alpha_{p13,2}+\alpha_{p14,2}+.....) \quad (1)$$

In (1), $\alpha_{L12}$ represent the link between knn1 and knn2. The directional property, $\alpha_{p11,2}$ gives information about source and destination pair of KNNs which are knn1 and knn2 in this case. The performance properties, $\alpha_{p12,2}$, $\alpha_{p13,2}$ and $\alpha_{p14,2}$, tell how the two KNNs, knn1 and knn2 are connected based on the knowledge domains. The system fades away the links that are used rarely and have an ability to recreate at a later time. The processor channel at the link manager helps in processing these link properties to create a link between knn1 and knn2.

## V. WHY ILS IS A KNOWLEDGE NETWORK

A knowledge network is a network formed by interlinking of knowledge components. Knowledge is nothing but collection of information which comprises of data and rules [11][12]. Knowledge network or as conventionally said Knowledge Based Systems (KBS) provide intelligent decisions with justifications provided by the rules [13]. Knowledge Bases like SenseNet models some important aspects of human reasoning in natural language but is built as a lexical knowledge base only [14].

ILS is a modified knowledge network where in each knowledge or a concept is represented by an entity called KNN. ILS is also a network formed by interlinking of these entities, which is the relationship between these entities. Thus in ILS, entities and the relationships between the entities makes knowledge. The intelligence to create the relationship between nodes lies within the entities itself. In ILS these relationships are the links. KNN is able to create links by virtue of its property to reason and infer the knowledge.

## VI. KNOWLEDGE EMBEDDING AND RETRIVAL

KNN being an important component of ILS, has a major role during knowledge embedding as well as during knowledge retrieval.

### A. Knowledge Embedding

When the information is embedded, ILS is just a network of cross linked nodes, where every concept or sub knowledge unit is a node and its multilateral links which are handled by Link Manager.

The role of KNN in embedding is as follows:

- The first quadrant of KNN, Input/ Output takes the input coming from the other nodes and passes it on to the next quadrant, storage and parser. As input it would require the knn id of the previous node and the remaining knowledge to be embedded. It is also responsible for passing the control over to the next KNN along with the information about the current node and knowledge to be embedded, once the linkage to the next knn is formed by other quadrants.
- When the control reaches to the second quadrant of KNN, it parses the knowledge to know the next KNN to link with. The parser then processes the link between the current and next KNN. Thus, the inbuilt parser at KNN is a linkbuilder, which means it helps in building links between nodes.
- Link manager plays a vital role in the creation of knowledge thread by connecting KNNs. It assigns the values to the link properties, which are processed by the parser, and save them with the link database which is the last quadrant of KNN. No two nodes can get arbitrarily linked, the performance properties decides the linkages between the nodes.

### B. Knowledge Retrieval

During knowledge retrieval, the processor channel of KNN analyzes the link properties and helps in creation of knowledge thread. Knowledge cone as shown in fig. 3 gets created as many sub-levels of knowledge are added under KNN. The sub-level size i.e. the no of nodes at a particular sub-level is proportional to the depth of the sub-level.

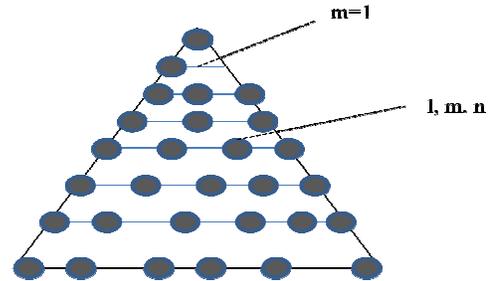

Figure 3  Knowledge Cone

*l - Knowledge Domain (KNN)*
*m –Knowledge level in domain KNN*
*n- Node at the mth knowledge level*

Nodes are not just words but are linked knowledge points that form a super cone which is an idea or knowledge. Knowledge retrieved can be thought of as cone of information formed by hierarchical expansion of nodes belonging to the domain of a knowledge component. Now since each KNN is part of a 3-D space as shown in fig. 3 each one is identified using 3-axis as (l, m, n) where each one of l, m, n can range from 1 to billions. Knowledge nodes linked at the mth level represent the knowledge threads that form the knowledge cone.

Each KNN is in 3-D domain (l, m, n). So when we move down the heap of knowledge, a dataset can be represented by first two dimensions (m,n). This dataset representation is part of 3-D representation of KNN i.e (l, m, n). This implies if a KNN is represented by (7, 3, 5) the dataset of sub-domain can be represented by (3, 5) as depicted in fig.4.

During knowledge retrieval all the performance properties are processed and now if the required property to the next KNN matches with the processed performance property the link is chosen. After this, the directional properties which include source and destination pair of KNNs are processed and the control is passed to next KNN. This process goes on till the node can process the link to the next node.

(,1)

(7,5,3)

Figure 4   Dataset Representation

To start a thread or to determine where to end the thread there needs to be necessary and sufficient condition e.g. African lions being the strong or not. To start we need to get the lions belonging to Africa. So the necessary condition is to check for the thread that starts from Africa and connects to lion at the sub-level of African knowledge component. From there the thread proceeds towards the properties of lion and if a natural link exists to the node strong, the thread stops there.

As the knowledge grows the link size also increases. From this we can derive the knowledge of ILS is integration of domain knowledge possessed by ILS.

$$K_{ILS} = \int K_d \qquad (2)$$

$K_{ILS}$ = Knowledge of ILS
$K_d$ = Domain Knowledge

## VII. SIMULATION AND EVALUATION

In our simulation, we embedded knowledge about what exist in the world. Embedding information resulted in creation of nodes and the multi-lateral links which represents the current knowledge of ILS.

Knowledge which is a stringed information cluster possessed by ILS is shown in fig.5. It represents the information to knowledge translation through linked structure. Knowledge cones are shown in fig.5 and the apex of a knowledge cone is represented by (l,0,0), where the first parameter, l, represent the knowledge domain.

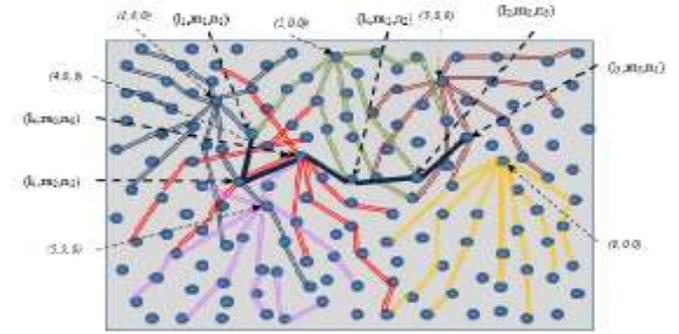

*Figure 5   Knowledge - A stringed information cluster*

The links are shown in fig.5 as bold connectors between nodes and each node is represented by a dataset $(l_i, m_j, n_k)$ which belongs to knowledge cone with $l_i$ as the apex.

To evaluate ILS, the evaluation criteria include number of links created during embedding and knowledge cones retrieved while retrieving knowledge for different concepts.

### A. Simulation Results

#### 1) Linkages during embedding

The new information that is embedded into existing ILS results in change in the structure of this knowledge network i.e., change in the number of information units, KNNs and the links in ILS. There can be numerous scenarios as shown in table 2, which can result while embedding information. One scenario may be where all nodes from the input information are already present leading to 0% addition of nodes into ILS but still result in adding up few more links whereas if all the nodes from the input information need to be created fresh, will have more number of linkages done, as depicted in fig. 5.

Table 2   ILS during embedding

| Scenario | Details | Nodes in knowledge | Nodes added (%) | links traversed | Links added (%) |
|---|---|---|---|---|---|
| Sc1 | All nodes present | 3 | 0.00 | 3 | 66.67 |
| Sc2 | Few nodes exist | 4 | 25.00 | 9 | 55.56 |
| Sc3 | Few nodes exist | 3 | 33.33 | 4 | 75.00 |
| Sc4 | Few nodes exist | 3 | 66.67 | 4 | 100.00 |
| Sc5 | All Nodes new | 4 | 100.00 | 6 | 100.00 |

During retrieval, the node(s) for which we want to retrieve knowledge threads is input to ILS. From the graph shown in fig. 6, we can deduce that embedding new knowledge which belongs to already existing knowledge heap, the number of knowledge units, KNNs to be added decreases.

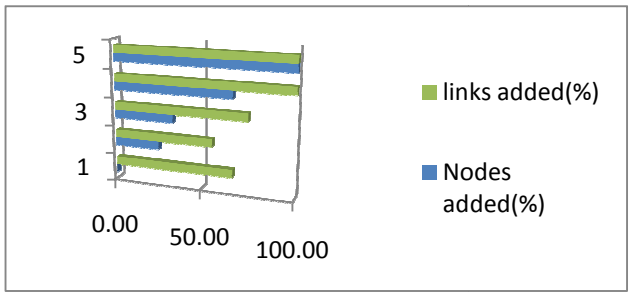

Figure 6 Linkages during embedding

This in turn leads to decrease in number of new links to be added but does not diminish it. So as the system embed more and more information, the more cross-linked knowledge network it becomes.

*2) Knowledge Cones during retrieval*

During retrieval, the knowledge unit(s) for which knowledge threads needs to be retrieved is input to ILS. For each unit that is present in the system, its KNN is identified and all the knowledge threads of that node are retrieved and a knowledge cone is formed.

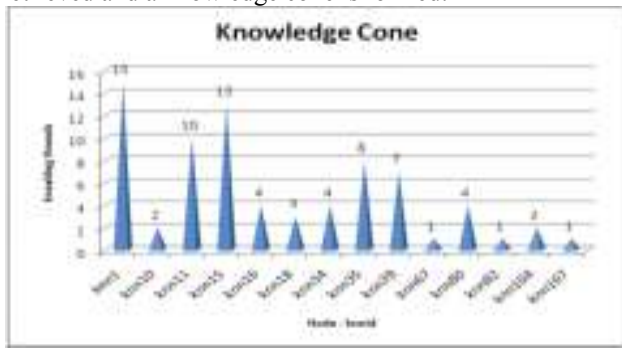

Figure 7 Knowledge Cones

The graph in fig. 7 depicts the knowledge cones retrieved for various nodes input to the developed system. The height of the cone is given by the number of knowledge threads extracted during retrieval. The height of the cones varies from one node to another which illustrate that the node which is linked to the highest degree will form a bigger cone whereas the nodes that are hardly ever linked will form a miniature cone.

## VIII. CONCLUSION

In this paper, we presented strategies which help in intelligent linking of information components through the link to form a network of knowledge and the strategy used to retrieve the knowledge embedded in ILS. During retrieval a knowledge cone is formed which helps in getting the related knowledge. The future work also includes validation of retrieval of much higher sizes of knowledge cones. ILS has been tested for limited knowledge of few domains till now. However, it is required to expand the information to larger dimensions to achieve practically viable knowledge units.